# Nonvisual Classification of Ground-Condition by Artificial Proprioception in an Amoeba-Inspired Autonomous Walking Robot

Hyoto Yamaguchi
RCIQE and Graduate School of IST
Hokkaido University
Sapporo, Japan
0009-0003-9057-1869

Zenji Yatabe
RCIQE and Graduate School of IST
Hokkaido University
Sapporo, Japan
0000-0003-2069-6677

Seiya Kasai
RCIQE and Graduate School of IST
Hokkaido University
Sapporo, Japan
0000-0003-1291-2587

***Abstract*—Nonvisual classification of ground condition based on a multimodal sensing approach was investigated for an amoeba-inspired autonomous walking robot. To classify ground condition without image sensing and processing, we implemented artificial proprioception by integrating a three-axis accelerometer, eight foot pressure sensors, and reservoir computing (RC). Even when large fluctuations in the sensor outputs are caused by dynamic motions of a four-legged robot in walking, our system can classify the ground condition, flat or rough, with high accuracy. We demonstrate on-site switching of walking gait depending on ground condition in the robot. We also discuss the contribution of each sensor to ground condition classification.**

## I. INTRODUCTION

Blind locomotion is an interesting issue for autonomous legged robots, in particular when it should work under low-visibility condition. Recently, legged robots can traverse challenging terrain using proprioceptive histories, contact information, or tactile sensing [1]–[3]. However, these approaches need large-scale simulation, extensive prior training, or substantial onboard computation [1]–[4]. Therefore, for small autonomous robots with limited energy, payload, and space, we have investigated applying amoeba-inspired solution search [5], [6] to a four-legged walking robot [7]. The robot walks by searching for a feasible motion and executing it step by step, called amoeba-inspired walking. This walking makes it possible to go over rough terrain with a commercially available small micro-controller, although the behavior is not organized and locomotion efficiency is low.

A considerable problem in our robot is to correctly perceive the ground condition by the proprioceptive sensing approach in press[8]. Acceleration, vibration, impact, foot pressure, and contact timing are affected by both robot motion and ground condition, allowing the robot to infer terrain without a camera [9], [10]. Previously we have investigated this approach based on the accelerations, however, it has been difficult to perceive the information necessary for blind locomotion, because the sensor signals are fluctuated severely due to large and irregular movement of the body attitude induced by nonperiodic walking gait. The irregular walking step cancels out the information of ground condition. This prevents our robot from making the appropriate behavioral decision in terms of blind locomotion.

On the above background, in this study, we investigate the multi-modal sensing-based artificial proprioception to perceive ground condition for the amoeba-inspired autonomous walking robot. Our previous work demonstrated the classification of ground conditions between flat and rough during normal walking based on three-axis accelerations using reservoir computing (RC) in press[8]. The remaining issue is perception during the amoeba-inspired walking. To improve the classification accuracy, we attempt to increase input information for RC by adding time-series data of pressure sensors on feet and also optimize the time series data length for classification. Then we demonstrate the on-site walking gait switching between normal and amoeba-inspired walks based on the classification of ground condition.

## II. SYSTEM DESIGN

Fig. 1(a) and 1(b) show photographs of an amoeba-inspired four-legged robot and the prepared experimental ground, respectively. For the controller of the robot, an Arduino Uno R4 Minima was used, which implemented an amoeba-inspired walking algorithm, a normal walking algorithm, RC models for ground condition classification, and a walking pattern switching mechanism, as shown in Fig. 2. The robot had a three-axis accelerometer (MMA8451), and eight foot-

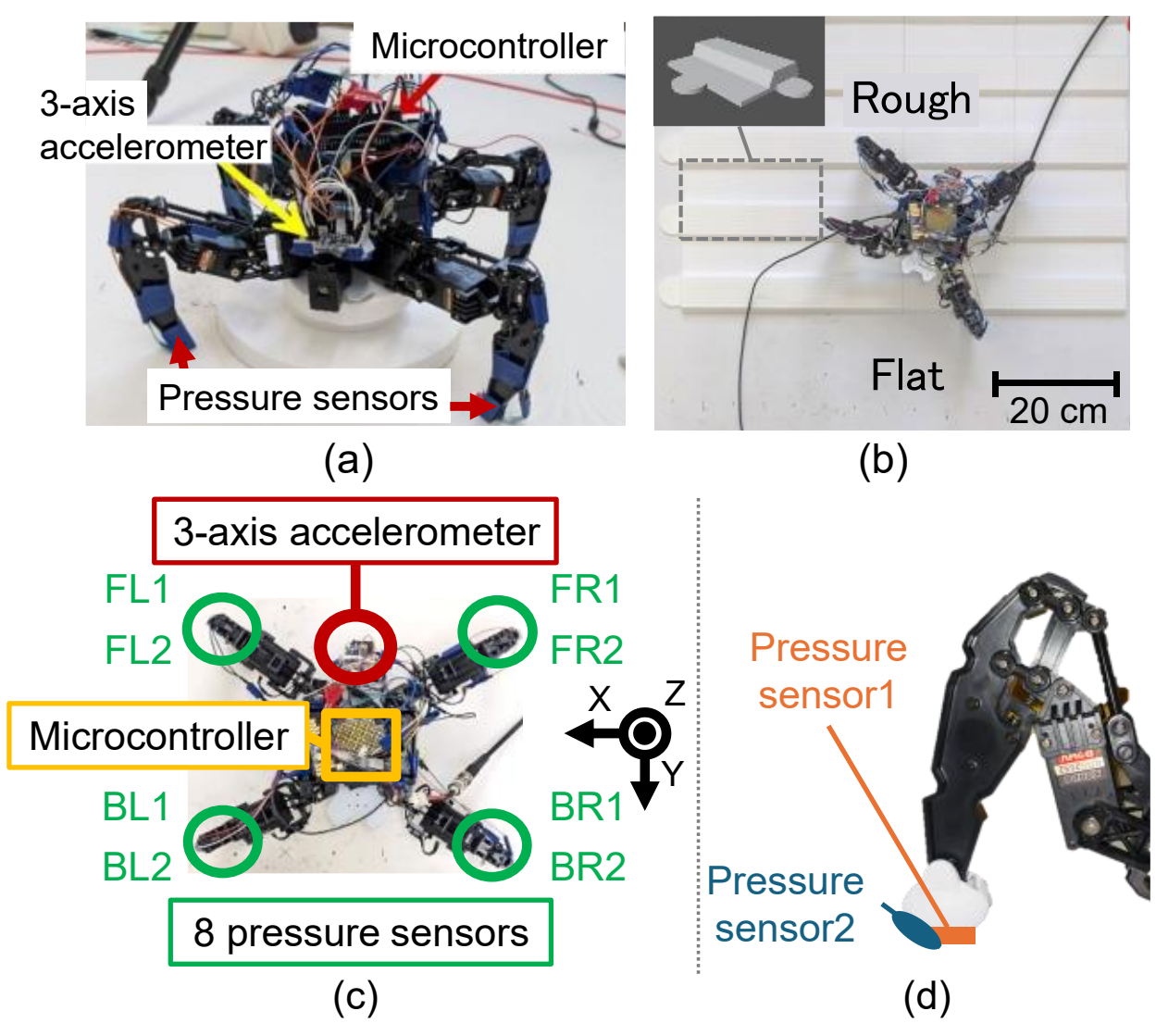

Fig. 1. (a) Amoeba-inspired four-legged walking robot, (b) top view of test ground, (c) layout of sensors on the robot, and (d) positions of foot pressure sensors.

This work is supported in part by JSPS Core-to-Core Program "Material Intelligence", JSPS KAKENHI Grant No. 26K01001, and by Grants for Revitalization of Regional Universities and Industries "Realization of a semiconductor complex base triggered by next-generation semiconductors and revitalization of local economies".

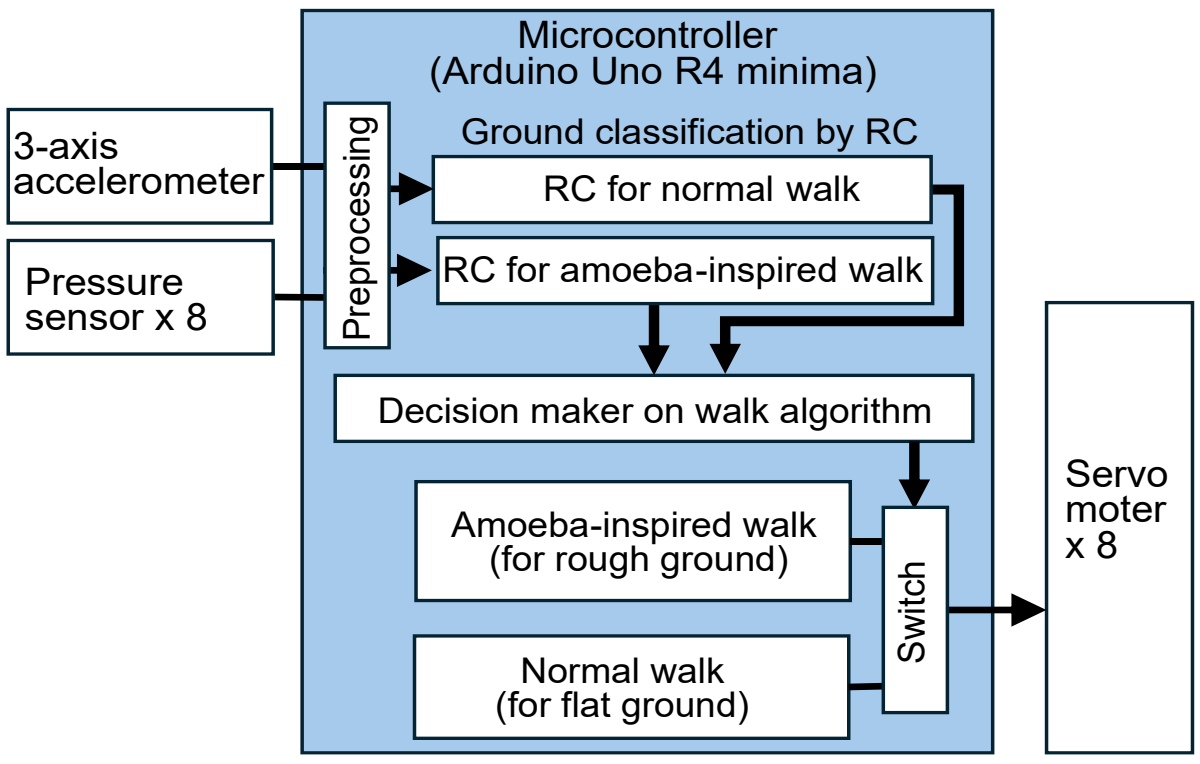

Fig. 2. Control system architecture of autonomous robot for switching walking gait depending on ground condition.

pressure sensors (MF01A-N-221-A04) for multimodal sensing as shown in Fig. 1(c). Each foot had two pressure sensors, because a single sensor missed detecting the ground contact depending on the angle of the leg contact (Fig. 1(d)). The robot had eight servo motors, corresponding to eight leg degrees of freedom. The amoeba-inspired solution-search mechanism in the microcontroller searched for a feasible leg state step by step in real time under constraints that avoided falling, retreating, and unstable postures [7].

The RC for amoeba walking developed in this study was trained to classify between flat and rough terrains during the amoeba-inspired walk. The system switched between two walking algorithms in accordance with a simple rule, where the normal walk was selected when the ground was flat and the amoeba-inspired walk was selected when the ground was rough. Both RC classifiers used an Echo State Network (ESN) which had a fixed recurrent neural network of nonlinear nodes, called reservoir layer, and a single trainable readout layer [11], [12].

For pretreatment of input data, sensor values were standardized using the mean and standard deviation of the training dataset. The smoothing was made on the RC output by a moving average, then the output was classified into flat or rough by thresholding. For on-site classification, the trained readout weights of the RC models and preprocessing coefficients were transferred to the RCs in the microcontroller on the robot.

## III. Experimental Method

### *A. Data Collection and Offline RC-model Training*

The rough terrain was prepared by arranging mesa blocks with heights of 10, 24, and 36 mm in arbitrary order. We obtained time-series data of 3-axis accelerations and eight pressure sensors during the robot walked on both flat and rough ground by the amoeba-inspired walk (Fig. 3). The RC for the amoeba-inspired walk was trained using the obtained data set. Then, we narrowed the data streams from eleven to three to maintain the ground condition classification accuracy as possible. This was because it was necessary to reduce the computational load for the microcontroller with the limited computing power. On the other hand, for normal walk, we reuse the RC trained in the previous study in press[8]. The input data for this RC included only 3-axis accelerations.

The numbers of walking steps obtained for training and testing of the RC were 646 steps on flat ground and 675 steps on rough ground, respectively. The obtained sensor data were divided at a ratio of 15:70:15, for using in optimization of hyperparameters, training, and validation test, respectively. To increase the classification accuracy, we accumulated the sensor data for several walking steps in classification: 1, 5, 10, 15, and 20 steps. Increasing the number of walking steps, the length of sensor history increased. The readout weights of the RC were trained on an external PC. Hyperparameters were optimized using Optuna [13]. The exploration ranges and optimized parameter values are summarized in Table I. To improve generalization, zero-mean Gaussian noise was added to the standardized inputs during training, and variation level of the noise was also optimized together with the ESN hyperparameters [14].

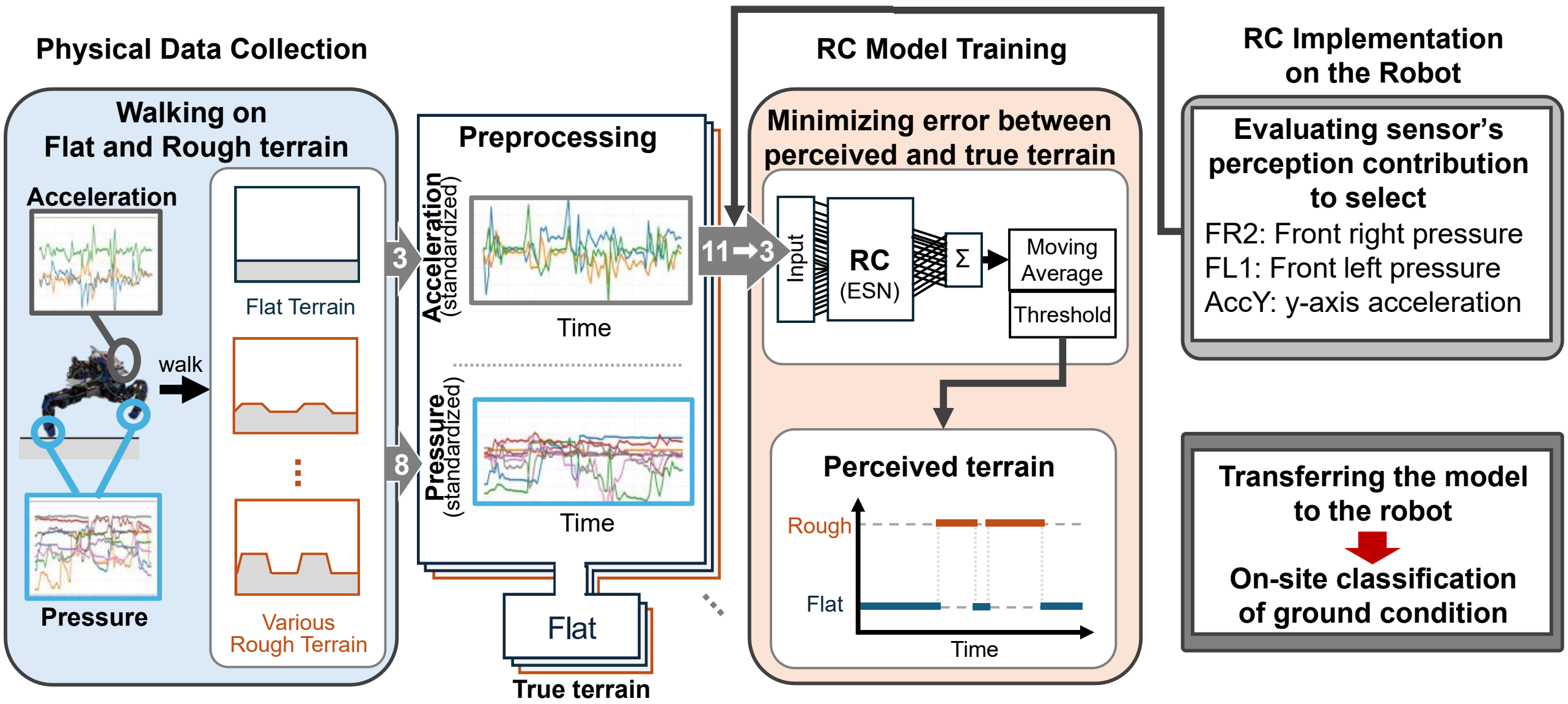

Fig. 3. Process of ground-condition classification in a four-legged amoeba-inspired autonomous walking robot.

TABLE I. HYPERPARAMETERS OF THE RC FOR AMOEBA-INSPIRED WALKING: EXPLORATION RANGE FOR OPTIMIZATION AND OPTIMIZED VALUES

| Parameter | Exploration range | Optimized |
|---|---|---|
| Nodes | 20-100 | 60 |
| Density | 0.01-0.25 | 0.13 |
| Input scale | 0.05-2.5 | 1.68 |
| ρ (Spectral radius) | 0.5-1.5 | 0.93 |
| Variation level | 0-0.3 | 0.25 |
| Average window | 5-300 | 274 |
| λ (Ridge coefficient) | $10^{-5}$ - $10^{2}$ | 0.012 |

### B. On-Site Classification and Gait Switching

The trained RCs were used to classify ground condition and to switch the walking gait on site. The robot was expected to move with normal walk on the flat ground and to move with the amoeba-inspired walk on the rough ground. To avoid misclassification of ground condition, switching of the walking algorithms was carried out based on different policies for the normal and amoeba-inspired walking. In the case of the normal walk, the system switched the algorithm to the amoeba-inspired walk when the RC for the normal walk continuously indicated rough terrain for a certain duration time. In the same way, in the case of the amoeba-inspired walk, the system switched the algorithm to the normal walk when the RC for the amoeba-inspired walk indicated flat continuously for a certain period. The period was determined based on the experimental result.

## IV. RESULTS

### A. Classification of Ground Condition

As shown in Fig. 4, the evaluated classification accuracy of the trained RCs for the amoeba-inspired walk using only acceleration data was 54 %, whereas that using all acceleration and pressure sensor data was 82 %. In both cases, classification was made at a single walking step. Multimodal sensing improved the classification accuracy. Fig. 5 shows the measured classification accuracies as a function of the number of walking steps for the amoeba-inspired walk with 3-axis acceleration data (referred to as ACC), and with all eleven sensor data (referred to as MM). In the case of ACC, the accuracy was around 55 % for both flat and rough terrains. The accuracy was not improved even though the walking step was increased. On the other hand, in the case of MM, the accuracy was obviously improved to 80 % on average at one walking step. The accuracy was further improved by increasing the number of steps and it achieved 97 % at 10 steps. It reached almost 100 % at 20 steps. Increasing the number of walking steps, response time was also increased. Thus, we considered that 10-step was feasible for classification in our robot.

Next, we narrowed down the input data for the RC to front-left-leg pressure sensor 1 (FL1), front-right-leg pressure sensor 2 (FR2), and $y$-axis acceleration (AccY). We experimentally found that these selected data could almost maintain the classification accuracy as shown in Table II.

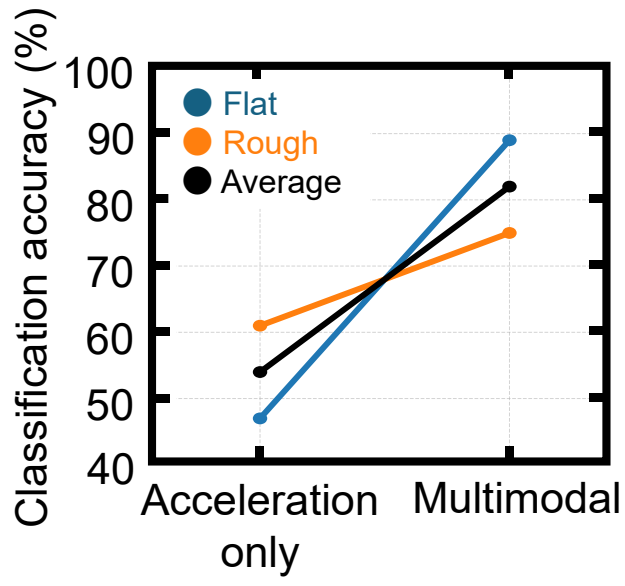

Fig. 4. Evaluated ground-condition classification accuracy.

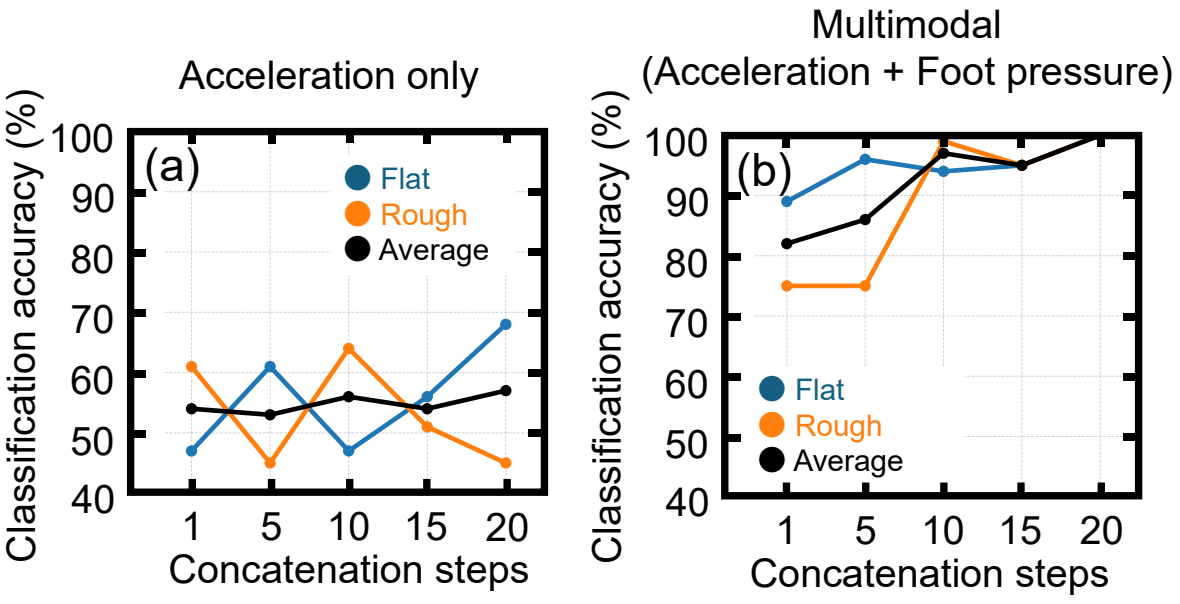

Fig. 5. Classification accuracy as a function of walking steps for classification in offline (a) using 3-axis accelerations and (b) 3-axis accelerations and 8-pressure sensors.

TABLE II. ACCURACY OF GROUND CONDITION CLASSIFICATION USING MULTIMODAL SENSING. ALL SENSORS INCLUDE 3-AXIS ACCELERATIONS AND 8 PRESSURE SENSORS

| | All sensors offline | FR2+FL1+AccY offline | FR2+FL1+AccY on-site |
|---|---|---|---|
| Flat | 94 % | 91 % | 70 % |
| Rough | 99 % | 81 % | 88 % |
| Average | 97 % | 86 % | 79 % |

### B. Gait Switching

For gait switching, it was necessary to adjust the timing of on-site algorithm switching from the amoeba-inspired walk to the normal walk. It was found that walking on rough ground often made a misclassification. From the observation of robot walking, the misclassification was found to continue for approximately 10 walking steps, 12 s in time. Therefore, we set the threshold of gait switching timing to 12 walking steps, 14.4 s in time. This threshold was applied when the robot transitioned from rough to flat ground.

Fig. 6 shows the snapshots of the robot walking from rough to flat ground. We examined two scenarios. As shown in Fig. 6(a), the robot started with the amoeba-inspired walk on rough ground, whereas as shown in Fig. 6(b) the robot started with the normal walk. In the first scenario, the robot performed the amoeba-inspired walk. After a while, the robot escaped from rough ground and reached flat ground. In this case, the robot needed to detect ground condition by the RC for the amoeba-inspired walk. After walking on flat ground for 120 s by the amoeba-inspired walk, the robot found the ground flat and switched the gait to the normal one. In the second scenario as shown in Fig. 6(b), the robot immediately found that ground was rough and it switched the gait from the normal to amoeba-inspired walk. After that, the robot walked and escaped from rough ground to flat. It took 760 s to find ground flat, suggesting that the system misclassified the ground condition many times. The observed behaviors of the robot demonstrated on-site non-visual perception of ground condition using a multimodal sensing approach, although the robot needed a rather long time for classification.

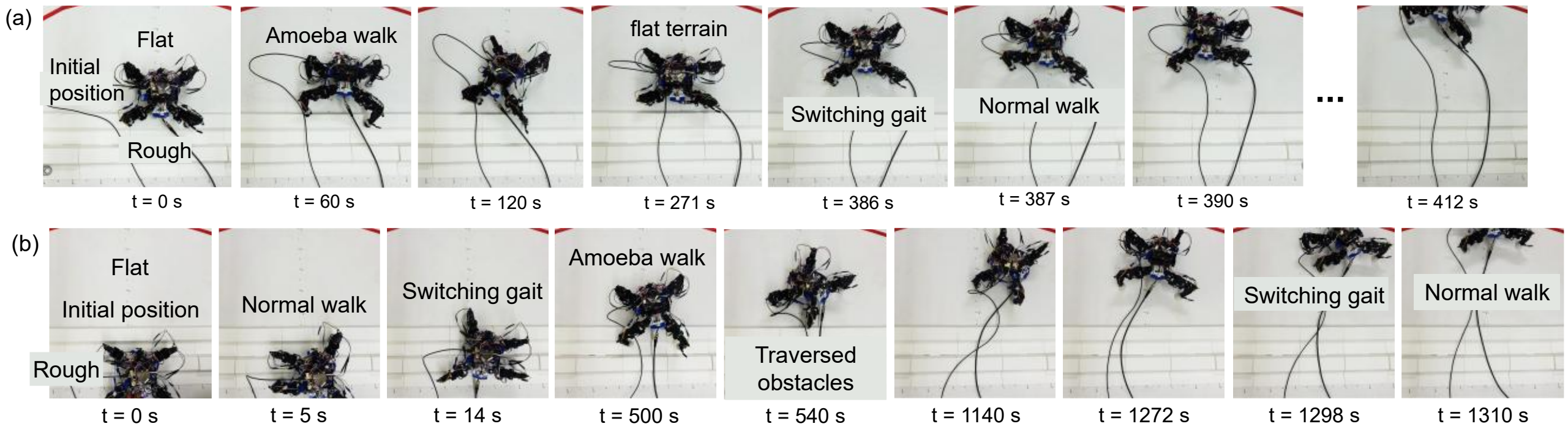

Fig. 6. Snapshots of robot walking with on-site walking gait switching depending on ground condition: (a) starting with amoeba-inspired walk (scenario 1) and (b) starting with normal walk (scenario 2).

## V. DISCUSSION

The multimodal sensing approach was found to improve classification accuracy for the amoeba-inspired autonomous walking robot in terms of ground condition. We analyzed which sensor information played an important role in the classification. Fig. 7(a) shows the evaluated contribution to classification of each sensor output to the RC output. It was found that the contributions of FR2 and FL1 were large, whereas the contributions of the accelerations were small. This result explains why high classification accuracy was achieved when the input of the RC was MM including FL1, FR2, and AccY data. The result also suggested that the pressure sensor data from the front leg was dominant on classification of ground condition, that is, the position of the foot pressure sensor was involved in classification. On the other hand, the contribution of the pressure sensors was bilaterally asymmetric even though the physical structure of the robot was symmetric.

To understand the large contribution of the specific sensor, the sensor sensitivity was analyzed. The result is shown in Fig. 7(b). Here, the sensor sensitivity was evaluated by discriminability using Cohen's method [15]. A clear correlation was observed between the contribution to the classification accuracy and the discriminability. The evaluated correlation coefficient $r$ was 0.87. The observed difference in the contribution shown in Fig. 7(a) was attributed not only to the mechanical role of each body part but also to variation in the discriminability, in particular pressure sensors. Bilateral contribution of the pressure sensors in the front legs shown in Fig. 7(a) might arise from variation in discriminability, although the standardization was made on the sensor data. Further analysis is necessary to understand the variation in contribution of the sensors.

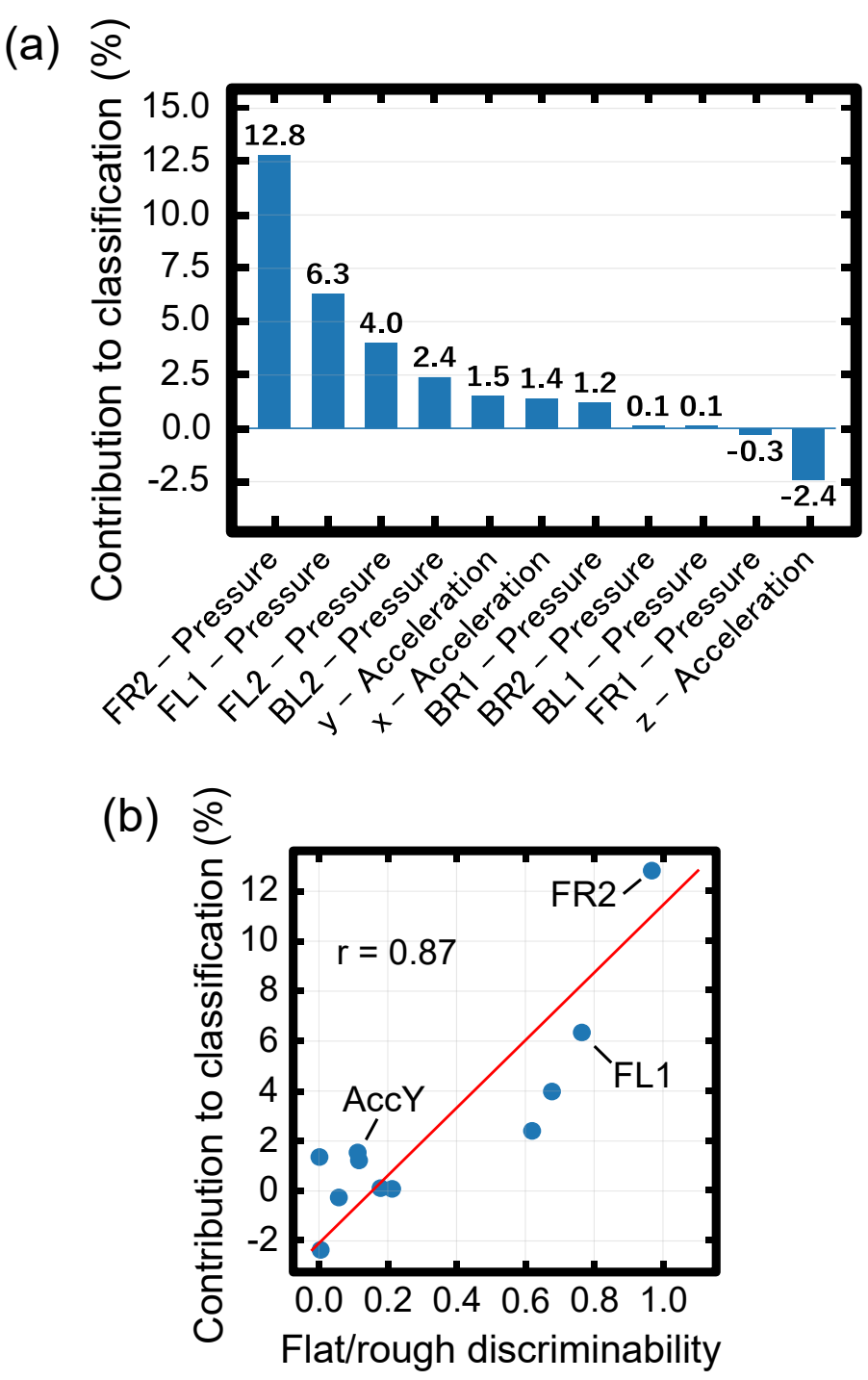

Fig. 7. (a) Evaluated contribution of each sensor to ground condition classification and (b) correlation between flat-rough discriminability among the sensors and contribution to ground-condition classification.

## VI. CONCLUSION

We investigated a nonvisual perception of ground condition in an amoeba-inspired autonomous walking robot based on the multimodal sensing approach utilizing reservoir computing (RC). By combining accelerations of the robot body and foot pressures, accurate classification between flat and rough ground was successfully achieved. We demonstrated on-site walking gait switching depending on ground condition in our amoeba-inspired walking robot installed our nonvisual classification system. The obtained result indicated the feasibility of our approach for the autonomous robot to obtain the surrounding information in a computationally compact manner without power-consuming image processing.